\documentclass[]{dataflow}

\usepackage{latexsym}
\usepackage[utf8]{inputenc}
\usepackage{graphicx}
\usepackage{booktabs}
\usepackage{multirow}
\usepackage{array}
\usepackage{colortbl}
\usepackage{amsmath}
\usepackage{pifont}
\usepackage{url}
\usepackage{fontawesome5}

\newcommand{\cmark}{\ding{51}}
\newcommand{\xmark}{\ding{55}}
\definecolor{guidanceblue}{RGB}{222,235,247}
\newcommand{\gcell}[1]{\cellcolor{guidanceblue}#1}

\title{WorkSurface-Bench: Benchmarking Enterprise Agents on Multi-Surface Knowledge Routing}

\author[1,2,*]{Hao Liang}
\author[1,*]{Meiyi Qiang}
\author[3,*]{Sizhe Qiu}
\author[3,*]{Linzhuang Sun}
\author[1,2]{Wentao Zhang}

\affiliation[1]{Peking University}
\affiliation[2]{Zhongguancun Academy}
\affiliation[3]{University of Chinese Academy of Sciences}

\abstract{
Enterprise agents often need several kinds of knowledge at once: documents
for narrative facts, tables for calculations, and dependency graphs for file
relationships. Existing benchmarks usually test retrieval or tool use, but do
not separate \emph{choosing the right knowledge form} from using it correctly.
We introduce \textbf{WorkSurface-Bench}, which evaluates this choice as
\emph{surface routing}. The benchmark projects five persona-scoped
Workspace-Bench-Lite workspaces onto document, table, and graph surfaces and
contains 1{,}151 atomic tasks. Its reference answers are auditable: table answers
come from executed DuckDB queries, document answers are checked against
verbatim spans, and graph answers derive from source annotations. We evaluate
four backbones under six agent settings, yielding 27{,}624 retained
trajectories with no protocol errors. Gold-constrained agents reach
98.7--99.8 Route F1, yet Answer remains 56.1--75.3\%. A matched intervention
shows that surface hints improve Answer for three of four models, whereas
removing irrelevant tools mainly improves routing and efficiency. In an
independent three-annotator audit, all 200
sampled tasks pass all six criteria by majority vote. We release the
construction pipeline, scoring code, and agent harness at
{\Urlmuskip=0mu plus 1mu\relax
\url{https://github.com/haolpku/WorkSurface-Bench}}.
}

\def\emailicon{\raisebox{-1.5pt}{\includegraphics[height=1.05em]{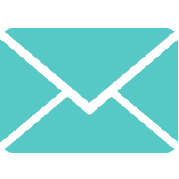}}}
\def\githubicon{\raisebox{-1.5pt}{\includegraphics[height=1.05em]{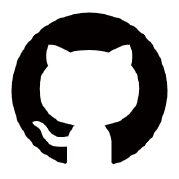}}}
\def\huggingfaceicon{\raisebox{-1.5pt}{\includegraphics[height=1.05em]{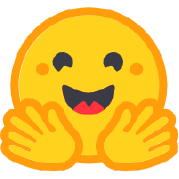}}}

\contribution[*]{Equal contribution}
\checkdata[ \emailicon \hspace{0.3em} Correspondence ]{\email{wentao.zhang@pku.edu.cn}}
\checkdata[ \githubicon \hspace{0.3em} Source Code ]{\href{https://github.com/haolpku/WorkSurface-Bench}{\texttt{haolpku/WorkSurface-Bench}}}
\checkdata[ \faGlobe \hspace{0.57em} Project Page ]{\url{https://haolpku.github.io/WorkSurface-Bench/}}
\checkdata[ \huggingfaceicon \hspace{0.3em} Dataset ]{\url{https://huggingface.co/datasets/lhpku20010120/WorkSurface-Bench}}

\begin{document}
\maketitle

\renewcommand{\thefootnote}{\arabic{footnote}}
\pagestyle{fancy}
\fancyhf{}
\fancyhead[L]{OpenDCAI Technical Report}
\fancyhead[R]{\thepage}

\newpage

\begin{figure}[t]
\centering
\includegraphics[width=0.98\textwidth]{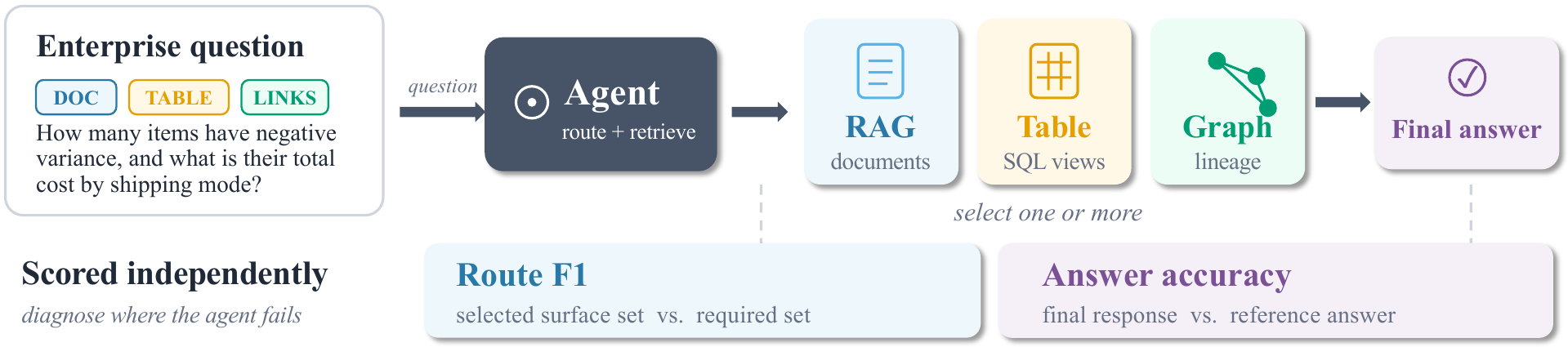}
\caption{WorkSurface-Bench projects a shared enterprise benchmark corpus onto three
canonical surfaces (RAG, Table, Graph) and scores \emph{Route} (surface
selection) and \emph{Answer} (final correctness) as separate metrics.}
\label{fig:teaser}
\end{figure}

\section{Introduction}
\label{sec:intro}

Consider an operations analyst asking: \emph{``Which source files feed the
negative-variance inventory report, and what is the total logistics cost by
shipping mode?''} The agent must read a report for context, query a table for
the total, and follow a dependency graph to identify the source files. These
are not interchangeable operations. Document retrieval cannot perform a SQL
aggregate, and a table query does not reveal file lineage. Before answering,
the agent must first decide which kinds of knowledge the question requires.

We call each kind of knowledge a \emph{surface}. In this paper, \textbf{surface
routing} means selecting the document, table, or graph surfaces needed for a
question. This capability is usually hidden inside an end-to-end score. RAG
benchmarks \citep{friel2024ragbench,chen2020hybridqa,chen2021ottqa} mainly ask
whether the final answer is correct. If a model chooses the wrong source or
retrieves the wrong evidence, both failures simply produce an incorrect
answer. Tool-use benchmarks
\citep{huang2023metatool,chen2023teval,li2023apibank} evaluate API selection,
but they do not test whether an agent recognizes that a question requires a
different representation of knowledge, such as SQL rather than prose search.
For deployment, this distinction matters: choosing the right surface and
reasoning correctly after that choice are separate abilities.

We introduce \textbf{WorkSurface-Bench} (Figure~\ref{fig:teaser}) to measure
both abilities. We project five persona-scoped workspaces onto three routable
surfaces: a document knowledge base, a DuckDB table registry, and a file
dependency graph. Procedural SOPs remain task metadata rather than a fourth
routable surface. Each task receives four scores---Route, Evidence, Answer,
and Efficiency---so the evaluation shows whether an agent chose the right
surface, found the right evidence, answered correctly, and used a reasonable
amount of computation. Reference answers never come directly from model free
text: we execute table queries, verify document spans, and derive graph answers
from source annotations.

This separation makes failures easier to interpret. A low Route score means
the agent chose the wrong surfaces; a low Evidence score means it chose the
right surfaces but did not access all required artifacts. High Route and
Evidence with low Answer is consistent with a downstream computation or
synthesis error, although the artifact-level Evidence metric does not prove
correct semantic use. End-to-end accuracy alone cannot expose these stages.

We evaluate four backbones (GPT-4o-mini, DeepSeek-V4-Pro, Gemini-3.1-Pro, and
GPT-5.5) under six agent settings on 1{,}151 tasks. The main result is simple:
\textbf{better routing does not guarantee a better answer}. Gold-constrained
agents reach 98.7--99.8 Route F1, while Answer remains 56.1--75.3\%. A matched
gold-hint condition further separates knowing the required surfaces from
removing irrelevant tools. Even after an agent knows where to look, it can
still retrieve incomplete evidence, combine sources incorrectly, or make a
computation error.

Our contributions are:
\begin{enumerate}
\item \textbf{A multi-surface routing benchmark}
(\S\ref{sec:dataset}): 1{,}151 tasks from five persona-scoped workspaces, projected
onto document, table, and graph surfaces. To our knowledge, it is the first
workspace benchmark to score surface routing separately from the Answer metric.
\item \textbf{A verifiable construction pipeline}
(\S\ref{sec:derivation}): deterministic rules and verified LLM-assisted
proposals produce gold answers from executed queries, checked spans, or source
graph annotations rather than model-generated values.
\item \textbf{A diagnostic scoring protocol} (\S\ref{sec:scoring}): Route,
Evidence, Answer, and Efficiency are scored separately, making it possible to
locate failures that one end-to-end score would hide.
\item \textbf{Evidence that routing and answering differ}
(\S\ref{sec:experiments}): gold-surface guidance improves Route F1 much more
than Answer across four complete backbone sweeps with no retained protocol errors.
\end{enumerate}

\section{Related Work}
\label{sec:related}

% ---- inlined from table_prior.tex ----
% Table 1: prior-benchmark comparison (paper_spec §3.6)
\begin{table}[t]
\centering
\small
\resizebox{\textwidth}{!}{%
\begin{tabular}{lccccc}
\toprule
\textbf{Benchmark} & \textbf{\#Surfaces} & \textbf{Route metric} & \textbf{Enterprise src} & \textbf{Skill/SOP} & \textbf{Graph type} \\
\midrule
Workspace-Bench 1.0~\citep{tang2026workspacebench} & files (1 form) & \xmark & \cmark & rubric & file-dep (1-hop) \\
STaRK~\citep{wu2024stark} & 2 (text+KG) & \xmark & \xmark & \xmark & concept KG \\
HybridQA~\citep{chen2020hybridqa} & 2 (text+table) & \xmark & \xmark & \xmark & --- \\
OTT-QA~\citep{chen2021ottqa} & 2 (text+table) & \xmark & \xmark & \xmark & --- \\
T\textsuperscript{2}-RAGBench~\citep{isgorur2026t2ragbench} & 2 (text+table) & \xmark & \xmark & \xmark & --- \\
MMQA~\citep{talmor2021mmqa} & 3 (text+table+image) & \xmark & \xmark & \xmark & --- \\
MetaTool~\citep{huang2023metatool} & tools (100s) & \cmark\,(tool) & \xmark & \xmark & --- \\
T-Eval~\citep{chen2023teval} & tools & \cmark\,(step) & \xmark & \xmark & --- \\
API-Bank~\citep{li2023apibank} & APIs ($\sim$50) & \cmark\,(API) & \xmark & \xmark & --- \\
RAGBench~\citep{friel2024ragbench} & 1 (docs) & \xmark & partial & \xmark & --- \\
BIRD / Spider~\citep{li2023bird,yu2018spider} & 1 (SQL) & \xmark & \xmark & \xmark & --- \\
TheAgentCompany~\citep{xu2024theagentcompany} & env (tools+web) & \xmark & \cmark & \xmark & --- \\
\midrule
\textbf{WorkSurface-Bench (ours)} & \textbf{3 routable + SOP metadata} & \cmark\,(surface) & \cmark & \cmark\,(SOP) & file lineage+enrich \\
\bottomrule
\end{tabular}%
}
\caption{Positioning against prior benchmarks. WorkSurface-Bench projects
three routable knowledge surfaces over a shared enterprise corpus, attaches
SOP metadata, and scores surface routing separately from answer correctness.
Entries are based on the cited benchmark descriptions.}
\label{tab:prior}
\end{table}
% ---- end table_prior.tex ----

WorkSurface-Bench sits at the intersection of four active benchmark
threads: multi-source retrieval-augmented QA, enterprise-scale agent
evaluation, tool and API routing, and semi-structured knowledge retrieval.
Table~\ref{tab:prior} places representative benchmarks along the two axes we
care about --- how many categorically distinct \emph{surfaces} are present,
and whether surface selection itself is a scored metric --- and shows that
the combination is not represented among the surveyed benchmarks.

\subsection{Multi-source Retrieval-Augmented QA}
The earliest attempts to move beyond single-surface RAG fused \emph{two}
sources into one integrated question. HybridQA \citep{chen2020hybridqa} pairs
Wikipedia tables with linked passages; OTT-QA \citep{chen2021ottqa} extends
this to 400K open-domain tables and 5M passages; T$^2$-RAGBench
\citep{isgorur2026t2ragbench} adds financial text-and-table retrieval; and
MMQA \citep{talmor2021mmqa} combines text, tables, and images. In these
benchmarks, the model is not scored for deciding which representation it
needs. mmRAG \citep{xu2025mmrag} is closest in covering text, tables, and
knowledge graphs and evaluating query routing, but converts them into
retrievable documents; we retain surface-native tools and score their actual
use. Enterprise Deep Search \citep{choubey2025deepsearch} evaluates multi-hop
retrieval across heterogeneous enterprise artifacts, whereas we separate
surface selection, artifact access, and answering over native document, SQL,
and lineage-graph interfaces.

\subsection{Semi-structured Knowledge Retrieval}
STaRK \citep{wu2024stark} evaluates retrieval over semi-structured knowledge
bases where a concept-level knowledge graph (entities, typed relations)
is blended with textual descriptions. Our Graph surface instead encodes
workspace-native file lineage (file-depends-on-file, file-supports-output,
task-requires-file), supporting questions such as which spreadsheets feed a
dashboard. It therefore tests a different structure from concept-level KG
retrieval.

\subsection{Tool and API Routing}
MetaTool \citep{huang2023metatool}, T-Eval \citep{chen2023teval}, API-Bank
\citep{li2023apibank}, and ToolLLM \citep{qin2023toolllm} evaluate tool or API
selection; AgentBoard \citep{ma2024agentboard} tracks multi-turn progress.
They route among operational endpoints or action sequences. We instead route
among information types: tools execute a route, while surfaces determine the
representation the agent must consult. This decision is complementary to
retriever routing: an agent may first choose a representation and then select
a retriever, query program, or API within it. This layering clarifies whether
a failure occurs before or after a surface-native interface has been selected,
which matters when comparing modular agent systems and multi-stage pipelines.

\subsection{Enterprise Workspace Agents}
Workspace-Bench \citep{tang2026workspacebench}, TheAgentCompany
\citep{xu2024theagentcompany}, WorkArena \citep{drouin2024workarena}, AppWorld
\citep{trivedi2024appworld}, and $\tau$-bench \citep{yao2024taubench} target
end-to-end workspace, web, or API success. We inherit Workspace-Bench's source
and provenance pipeline, but isolate knowledge routing and expose file
dependencies as a lineage-query surface.

\section{Benchmark Construction}
\label{sec:dataset}

\begin{figure}[t]
\centering
\includegraphics[width=0.98\textwidth]{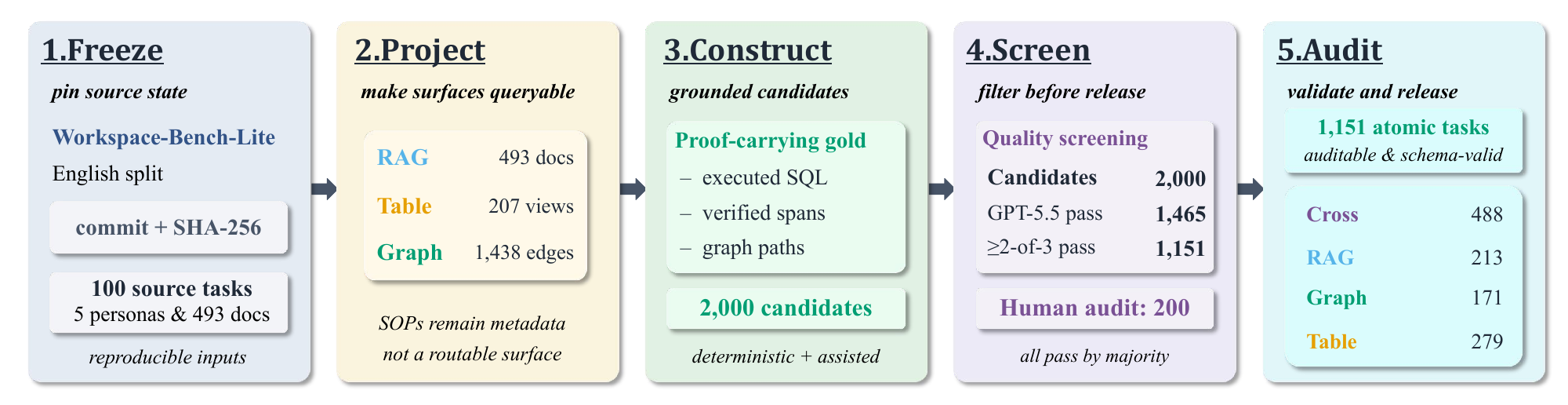}
\caption{Construction pipeline. Gold answers trace to executed DuckDB results,
verified document spans, or source dependency annotations; model-generated
proposals are grounded through these proof artifacts. Details in
\S\ref{sec:derivation}.}
\label{fig:pipeline}
\end{figure}

WorkSurface-Bench is built by a deterministic pipeline that adds no synthetic
facts, followed by a verifiable LLM-assisted expansion. Figure~\ref{fig:pipeline}
gives the overview; we detail each stage below and report the resulting
statistics in Tables~\ref{tab:stats-composition}
and~\ref{tab:stats-verification}.

% ---- inlined from table_stats.tex ----
\begin{table}[t]
\begin{minipage}[t]{0.47\textwidth}
\centering
\small
\renewcommand{\arraystretch}{1.08}
\begin{tabular*}{\linewidth}{@{\extracolsep{\fill}}lr@{}}
\toprule
\textbf{Dataset statistic} & \textbf{Value} \\
\midrule
Personas & 5 \\
Source tasks (WSB-Lite) & 100 \\
Atomic tasks & 1{,}151 \\
Task types & 4 \\
\quad cross\_surface & 488 \\
\quad rag\_only & 213 \\
\quad graph\_only & 171 \\
\quad table\_only & 279 \\
Persona count range & 39--595 \\
\bottomrule
\end{tabular*}
\caption{Dataset composition after verification and screening.}
\label{tab:stats-composition}
\end{minipage}
\hfill
\begin{minipage}[t]{0.49\textwidth}
\centering
\small
\renewcommand{\arraystretch}{1.08}
\begin{tabular*}{\linewidth}{@{\extracolsep{\fill}}lr@{}}
\toprule
\textbf{Artifact / check} & \textbf{Value} \\
\midrule
Routable surfaces & 3 \\
KB canonical docs & 493 \\
Table views (DuckDB) & 207 \\
Graph edges (total) & 1{,}438 \\
Executed table-bearing gold & 453/453 \\
RAG-bearing span verification & 601/601 \\
Graph-bearing path verification & 600/600 \\
2-of-3 screening pass & 1{,}151/2{,}000 \\
Efficiency budgets & 1{,}151/1{,}151 \\
\bottomrule
\end{tabular*}
\caption{Surface artifacts and executable verification coverage.}
\label{tab:stats-verification}
\end{minipage}
\end{table}
% ---- end table_stats.tex ----

\subsection{Source Data and Provenance}
\label{sec:provenance}
We derive the benchmark from the Workspace-Bench-Lite English split. Its data
pipeline is hybrid: task scenarios and dependency graphs are human-authored
and expert-validated against Lark/ByteDance workflows, while workspace files
combine public web resources with grounded LLM-generated artifacts. The result
is an enterprise-workflow benchmark with human-authored task, rubric, and graph
supervision. For reproducibility, the first pipeline stage records the source
commit and every input file's SHA-256 in a lock file. We assess closed-book
answerability separately (\S\ref{sec:closedbook}).

\subsection{Surface Projection}
\label{sec:surfaces}
Each persona workspace is projected onto three routable surfaces.
\textbf{RAG}: documents, reports, PDFs, and slide text are converted to
canonical UTF-8 markdown so routing is evaluated separately from OCR/parsing.
\textbf{Table}: normalized CSV/XLSX sheets are registered as per-workbook
DuckDB views with provenance columns; a coverage gate retains non-trivial
sheets (at least 30 rows, four columns with at least three rows, or five
distinct non-empty values in one column) as tables and maps the remainder to
RAG. \textbf{Graph}: the Workspace-Bench file
dependency graph enriched with programmatic edges (\texttt{mentions},
\texttt{schema\_overlap}, \texttt{version\_of}, \texttt{shared\_artifact}).
Repeated rubric patterns yield \texttt{applicable\_skills} metadata, which
serves as auxiliary task context rather than a routable surface. Each surface
passes executable eligibility and leakage checks reported in
Appendix~\ref{sec:appendix}.

\subsection{Task Derivation}
\label{sec:derivation}
Atomic tasks are derived in two passes. The \emph{deterministic} pass derives
golds directly from source artifacts: numeric
rubric values ($\$$1{,}710{,}971.47, 16.89\%, ``1000 records'') become gold
answers; one \texttt{graph\_only} task per source task asks which files are
required inputs; and \texttt{cross\_surface} tasks are emitted when an executed
DuckDB aggregate over a named column reproduces the rubric value. In the
\emph{verifiable LLM-assisted} pass, the model proposes questions together with
proof artifacts. Table and cross-surface proposals include a DuckDB query whose
executed result becomes the gold; RAG proposals include an answer verified
verbatim in the source document. Proposals with query errors or missing spans
are discarded. Graph-only golds come directly from source dependency
annotations. A final QC pass deduplicates the tasks and re-validates the schema.

\subsection{Automatic Validation and Model-assisted Screening}
\label{sec:qc}

We generate a 2{,}000-item candidate pool and first apply deterministic schema,
uniqueness, language, leakage, query-execution, span, and graph-path checks.
Every table-bearing item must reproduce its gold through a read-only query,
every RAG-bearing item must point to a verified span, and every graph-bearing
item must match a source dependency path. IDs and normalized questions must
also be unique.

We then apply a semantic screen for answerability, gold correctness,
naturalness, atomicity and unambiguity, necessity of the declared surfaces,
and absence of leakage cues. GPT-5.5 retains 1{,}465 candidates under this
strict rubric. Independent DeepSeek and Gemini audits produce the final
1{,}151-item set by requiring a strict pass from at least two of the three
auditors; 429 items pass all three. A strict pass requires an affirmative
judgment on every dimension, with failed or ``unsure'' judgments treated as
non-passes.

\subsection{Independent Human Audit}
\label{sec:human-eval}
Three annotators independently audit the same stratified 200-item sample using
the released evidence package and rubric. The sample covers task type,
persona, and answer type, including all 15 tri-surface items. All 200 items
receive majority-pass judgments on all six criteria. Surface necessity and
atomicity are unanimously positive for every item, and 192 items receive
unanimous judgments across all six criteria. We retain the item-level votes
and compute these rates on the originally sampled items.

\subsection{Dataset Distribution and Coverage}
\label{sec:distribution-coverage}

The final distribution reflects the candidates that satisfy verification and
surface-necessity criteria. It contains 488 cross-surface tasks:
314 RAG+Graph, 100 Graph+Table, 59 RAG+Table, and 15 requiring all three
surfaces. Tri-surface tasks are less frequent because each of the three
evidence forms must be independently necessary and the composed gold must
remain executable or traceable.

Persona frequencies follow the source workspaces and remain concentrated in
operational roles (Table~\ref{tab:stats-composition}).

\begin{table}[t]
\centering\small
\renewcommand{\arraystretch}{0.98}
\setlength{\tabcolsep}{4.2pt}
\begin{tabular*}{\textwidth}{@{\extracolsep{\fill}}clccccccc}
\toprule
\multicolumn{2}{c}{\textbf{Configuration}} & \multicolumn{3}{c}{\textbf{Routing}} & \multicolumn{2}{c}{\textbf{Agent performance}} & \textbf{Resource use} & \textbf{Overall} \\
\cmidrule(lr){3-5}\cmidrule(lr){6-7}
\textbf{Model} & \textbf{Setting} & \textbf{P} & \textbf{R} & \textbf{F1} & \textbf{Evidence} & \textbf{Answer} & \textbf{Eff.} & \textbf{Agg.} \\
\midrule
\multirow{6}{*}{GPT-4o-mini} & No-tool & -- & -- & -- & 0.0 & 0.0 & \textbf{98.6} & 9.9 \\
 & Always-RAG & 52.2 & 35.1 & 40.8 & 14.1 & 18.8 & 90.8 & 30.1 \\
 & Naive-router & 43.5 & 29.2 & 33.9 & 27.6 & 12.1 & 74.6 & 28.4 \\
 & ReAct-all & 66.8 & 79.4 & 70.4 & 59.7 & 48.7 & 53.5 & 57.9 \\
 & \gcell{Gold-constrained} & \gcell{\textbf{99.8}} & \gcell{\textbf{98.1}} & \gcell{\textbf{98.7}} & \gcell{\textbf{66.7}} & \gcell{56.1} & \gcell{75.0} & \gcell{\textbf{71.8}} \\
 & \gcell{Gold-hint/all} & \gcell{81.6} & \gcell{90.0} & \gcell{84.0} & \gcell{64.4} & \gcell{\textbf{58.6}} & \gcell{52.7} & \gcell{66.1} \\
\midrule
\multirow{6}{*}{DeepSeek-V4-Pro} & No-tool & -- & -- & -- & 0.0 & 0.0 & \textbf{96.1} & 9.6 \\
 & Always-RAG & 52.2 & 35.1 & 40.8 & 14.1 & 20.3 & 85.7 & 30.1 \\
 & Naive-router & 55.4 & 46.7 & 49.5 & 36.3 & 40.6 & 51.8 & 42.6 \\
 & ReAct-all & 69.0 & 99.3 & 79.1 & 77.2 & 67.9 & 27.2 & 69.4 \\
 & \gcell{Gold-constrained} & \gcell{\textbf{100.0}} & \gcell{\textbf{99.5}} & \gcell{\textbf{99.7}} & \gcell{77.0} & \gcell{\textbf{70.7}} & \gcell{43.4} & \gcell{\textbf{77.1}} \\
 & \gcell{Gold-hint/all} & \gcell{71.6} & \gcell{99.2} & \gcell{81.0} & \gcell{\textbf{77.5}} & \gcell{65.7} & \gcell{26.5} & \gcell{69.1} \\
\midrule
\multirow{6}{*}{Gemini-3.1-Pro} & No-tool & -- & -- & -- & 0.0 & 0.0 & \textbf{93.6} & 9.4 \\
 & Always-RAG & 52.2 & 35.1 & 40.8 & 14.1 & 18.8 & 83.3 & 29.3 \\
 & Naive-router & 79.6 & 63.9 & 69.1 & 48.9 & 60.5 & 39.6 & 57.1 \\
 & ReAct-all & 90.3 & 99.6 & 93.5 & 77.6 & 76.0 & 32.1 & 76.5 \\
 & \gcell{Gold-constrained} & \gcell{\textbf{100.0}} & \gcell{99.7} & \gcell{\textbf{99.8}} & \gcell{\textbf{78.5}} & \gcell{75.3} & \gcell{40.1} & \gcell{\textbf{78.9}} \\
 & \gcell{Gold-hint/all} & \gcell{96.9} & \gcell{\textbf{100.0}} & \gcell{98.0} & \gcell{78.3} & \gcell{\textbf{78.8}} & \gcell{31.3} & \gcell{78.7} \\
\midrule
\multirow{6}{*}{GPT-5.5} & No-tool & -- & -- & -- & 0.0 & 2.4 & \textbf{94.8} & 10.3 \\
 & Always-RAG & 52.2 & 35.1 & 40.8 & 14.1 & 22.7 & 85.9 & 30.9 \\
 & Naive-router & 79.5 & 63.4 & 68.7 & 46.4 & 55.3 & 44.2 & 54.9 \\
 & ReAct-all & 74.3 & 99.1 & 82.4 & 76.1 & 63.8 & 35.6 & 69.3 \\
 & \gcell{Gold-constrained} & \gcell{\textbf{100.0}} & \gcell{99.4} & \gcell{\textbf{99.6}} & \gcell{75.4} & \gcell{61.9} & \gcell{42.3} & \gcell{\textbf{73.4}} \\
 & \gcell{Gold-hint/all} & \gcell{79.1} & \gcell{\textbf{99.9}} & \gcell{86.1} & \gcell{\textbf{76.6}} & \gcell{\textbf{68.1}} & \gcell{32.1} & \gcell{71.5} \\
\bottomrule
\end{tabular*}
\caption{Main results (\%) across four models, six settings, and 1{,}151 tasks.
Route-F1, Evidence, Answer, and Efficiency form Agg.\ as defined in
\S\ref{sec:scoring}. No-tool Route is unavailable and scored zero. Shading
marks the two gold-surface conditions; bold marks the within-model best. All
27{,}624 retained trajectories are protocol-error-free.}
\label{tab:main}
\end{table}
\subsection{Dataset Statistics}
\label{sec:stats}
The benchmark comprises 1{,}151 atomic tasks over five personas
(Table~\ref{tab:stats-composition}): 488 \texttt{cross\_surface},
213 \texttt{rag\_only},
171 \texttt{graph\_only}, and 279 \texttt{table\_only}. Each task carries a
natural-language question, gold answer, required surfaces, expected tool types,
evidence annotations, and Workspace-Bench provenance. Answer-bearing golds
are auditable: all 453 table-bearing queries reproduce their reference
answers, all 601 RAG-bearing items match verified document spans, and all 600
graph-bearing items pass executable node/path checks
(Table~\ref{tab:stats-verification}).

\begin{table}[t]
\centering\small
\renewcommand{\arraystretch}{0.98}
\setlength{\tabcolsep}{4.2pt}
\begin{tabular*}{\textwidth}{@{\extracolsep{\fill}}clcccc}
\toprule
\multicolumn{2}{c}{\textbf{Configuration}} & \multicolumn{4}{c}{\textbf{Answer score by task type}} \\
\cmidrule(lr){3-6}
\textbf{Model} & \textbf{Setting} & \textbf{RAG} & \textbf{Table} & \textbf{Graph} & \textbf{Cross} \\
\midrule
\multirow{6}{*}{GPT-4o-mini} & No-tool & 0.0 & 0.0 & 0.0 & 0.0 \\
 & Always-RAG & \textbf{70.0} & 0.0 & 26.4 & 4.5 \\
 & Naive-router & 4.7 & 8.6 & 59.0 & 0.8 \\
 & ReAct-all & 19.7 & \textbf{88.5} & 67.1 & 32.2 \\
 & \gcell{Gold-constrained} & \gcell{58.7} & \gcell{86.0} & \gcell{\textbf{72.9}} & \gcell{32.0} \\
 & \gcell{Gold-hint/all} & \gcell{46.0} & \gcell{88.4} & \gcell{70.5} & \gcell{\textbf{42.8}} \\
\midrule
\multirow{6}{*}{DeepSeek-V4-Pro} & No-tool & 0.0 & 0.0 & 0.0 & 0.0 \\
 & Always-RAG & \textbf{80.8} & 0.4 & 24.1 & 4.1 \\
 & Naive-router & 74.2 & 74.5 & 20.5 & 13.5 \\
 & ReAct-all & \textbf{80.8} & 85.2 & 72.3 & 50.8 \\
 & \gcell{Gold-constrained} & \gcell{80.3} & \gcell{\textbf{91.0}} & \gcell{70.7} & \gcell{\textbf{54.9}} \\
 & \gcell{Gold-hint/all} & \gcell{78.9} & \gcell{86.7} & \gcell{\textbf{72.8}} & \gcell{45.5} \\
\midrule
\multirow{6}{*}{Gemini-3.1-Pro} & No-tool & 0.0 & 0.0 & 0.0 & 0.0 \\
 & Always-RAG & 81.2 & 0.0 & 25.6 & 0.0 \\
 & Naive-router & 82.2 & 87.4 & 47.1 & 40.4 \\
 & ReAct-all & 83.1 & 86.5 & 64.9 & 70.9 \\
 & \gcell{Gold-constrained} & \gcell{82.6} & \gcell{87.0} & \gcell{\textbf{68.1}} & \gcell{68.0} \\
 & \gcell{Gold-hint/all} & \gcell{\textbf{83.6}} & \gcell{\textbf{88.8}} & \gcell{65.3} & \gcell{\textbf{75.8}} \\
\midrule
\multirow{6}{*}{GPT-5.5} & No-tool & 12.7 & 0.0 & 0.0 & 0.0 \\
 & Always-RAG & 82.2 & 0.0 & 29.0 & 7.6 \\
 & Naive-router & 82.6 & 84.4 & 39.9 & 32.2 \\
 & ReAct-all & 83.1 & 86.2 & \textbf{73.9} & 38.9 \\
 & \gcell{Gold-constrained} & \gcell{83.6} & \gcell{85.5} & \gcell{\textbf{73.9}} & \gcell{34.6} \\
 & \gcell{Gold-hint/all} & \gcell{\textbf{84.0}} & \gcell{\textbf{87.6}} & \gcell{73.4} & \gcell{\textbf{48.2}} \\
\bottomrule
\end{tabular*}
\caption{Mean Answer (\%) by task type. Shading marks the two gold-surface
conditions; bold marks the within-model best.}
\label{tab:persurface}
\end{table}
\section{Scoring Protocol}
\label{sec:scoring}

For each task $i$, the evaluator consumes the final answer and tool trace and
returns four scores in $[0,1]$. Table~\ref{tab:main} reports macro-averages
over all 1{,}151 tasks. The decomposition separates surface selection, evidence
retrieval, answer correctness, and inference cost.

\paragraph{Route.}
Let $G_i\subseteq\{\textsc{rag},\textsc{table},\textsc{graph}\}$ be the gold
surfaces and $C_i$ the surfaces actually used in the trace. We compute
$P_i=|C_i\cap G_i|/|C_i|$, $R_i=|C_i\cap G_i|/|G_i|$, and their harmonic mean
$F^{\mathrm{route}}_i$. Empty predictions receive zero when a surface is
required. Table~\ref{tab:main} reports Route-P, Route-R, and Route-F1, but only
Route-F1 enters the aggregate. For No-tool, the Route columns are displayed as
``--'' because no routing action is available, while Route-F1 is scored as
zero whenever a surface is required.

\paragraph{Evidence.}
Each gold evidence item is matched against the trace: a RAG item is hit when
its source file is read, a table item when its registered view is queried, and
a graph path item only when its terminal evidence node is returned. For graph
items annotated as complete neighbor sets, every verified member must be
returned. The Evidence score $E_i$ is the fraction of matched artifact-level
items, so tasks with multiple annotations receive proportional rather than
surface-level credit. It measures access to the required artifacts, not
whether the final synthesis uses their contents correctly.

\paragraph{Answer.}
The Answer score $A_i$ depends on answer type. Integers with absolute value at
most 100 are matched exactly; other numbers use a 5\% relative-error tolerance.
Lists receive order-invariant set F1, strings and booleans use normalized exact
match, and abstention requires \texttt{INSUFFICIENT\_EVIDENCE}. Ordered
composite strings are compared element-wise, making the canonical
\texttt{item; value} form equivalent to a JSON array containing the same
elements. The dataset has no free-form items, so no model-based judge is used.

\begin{figure}[t]
\centering
\includegraphics[width=0.98\textwidth]{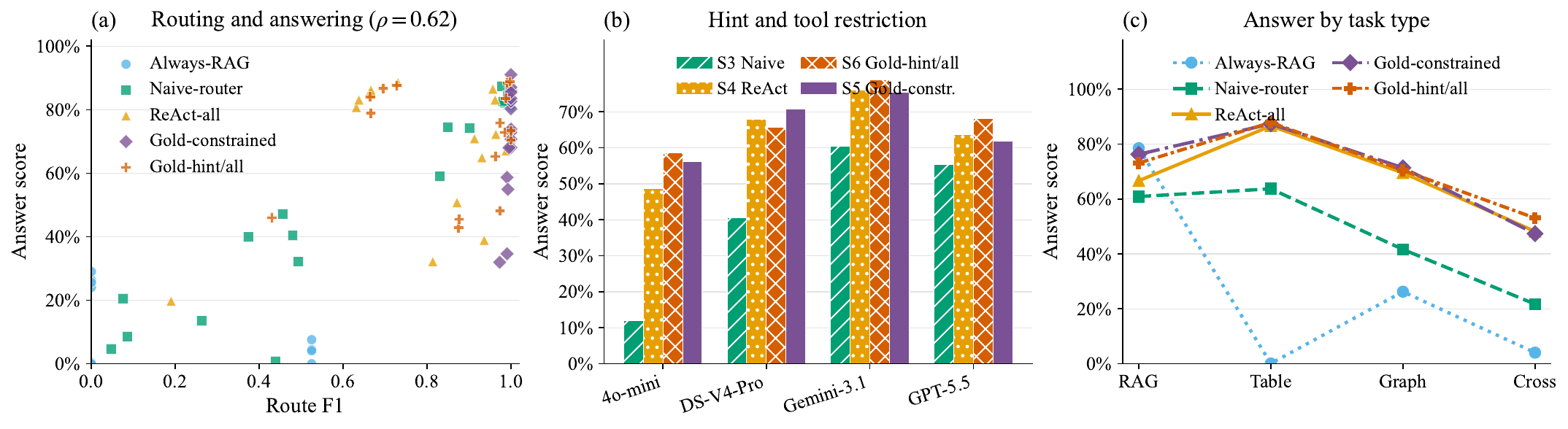}
\caption{Routing and answering exhibit related but distinct behavior.
(a) Route F1 versus Answer across 80 model--setting--task-type buckets
($\rho=0.62$). (b) Mean Answer for settings that vary routing information and
tool access. (c) Mean Answer by task type and setting.}
\label{fig:main}
\end{figure}
\paragraph{Efficiency.}
Let $t_i$ be the observed token count and $b_i$ the task-specific common
budget, fixed at twice the token count of the canonical GPT-4o-mini S5 trace.
The continuous linear score is
\[
Q_i=\max\!\left\{0,\,1-\frac{t_i}{2b_i}\right\}.
\]
Thus, $Q_i$ decreases from 1 at zero tokens to 0.5 at the common budget and
reaches 0 at twice the budget. If no positive budget is available, we assign the
neutral value $Q_i=1$; this defensive fallback is not triggered by the
released set.

\paragraph{Aggregate.}
The released set has no activated safety-threat annotations, so Safety is
omitted and does not enter this aggregate. We define the aggregate using four
evaluation hyperparameters:
\[
\begin{aligned}
\mathrm{Agg}_i={}&\alpha A_i+\beta E_i\\
                 &+\gamma F^{\mathrm{route}}_i+\delta Q_i,\\
\alpha+\beta+\gamma+\delta={}&1.
\end{aligned}
\]
Here $\alpha$, $\beta$, $\gamma$, and $\delta$ weight Answer, Evidence, Route,
and Efficiency, respectively. We set $\alpha=0.35$, $\beta=0.30$,
$\gamma=0.25$, and $\delta=0.10$: end-task correctness receives the largest
weight, evidence and routing remain substantial, and efficiency is secondary.
Equal weighting would assign 25\% to Efficiency and allow a cheap but
incorrect no-tool system to receive substantial credit. The 10\% weight keeps
efficiency consequential without allowing it to dominate task success. We
average $\mathrm{Agg}_i$ across tasks rather than reconstructing it from
rounded column means; component scores remain the primary diagnostic outcomes.

\section{Experiments}
\label{sec:experiments}

\subsection{Experimental Setup}
\label{sec:setup}

\paragraph{Backbones.}
Using identical prompts and tool schemas, we query four proxy-exposed API
identifiers: \texttt{gpt-4o-mini} \citep{openai2024gpt4omini},
\texttt{deepseek-v4-pro} \citep{deepseek2026v4},
\texttt{gemini-3.1-pro-preview} \citep{google2026gemini31pro}, and
\texttt{gpt-5.5} \citep{openai2026gpt55}. Released traces record the requested
identifiers (costs in Appendix~\ref{sec:reproducibility}).

\paragraph{Agent settings.}
Six settings span the routing autonomy spectrum:
(\textbf{S1}) \emph{No-tool} answers from parametric memory only ---
the closed-book baseline;
(\textbf{S2}) \emph{Always-RAG} forces a single \texttt{kb\_search} on every
task;
(\textbf{S3}) \emph{Naive-router} picks exactly one surface up front and
uses only its tools;
(\textbf{S4}) \emph{ReAct-all} exposes all three surfaces and lets the agent
freely interleave them; and
(\textbf{S5}) \emph{Gold-constrained} is given the required surfaces and can
only use tools within them; and
(\textbf{S6}) \emph{Gold-hint/all} receives the same surface names while all
three tool families remain available. Both conditions leave tool choice,
evidence retrieval, and answering to the model. S4$\to$S6 isolates the effect
of surface information, while S6$\to$S5 isolates the effect of removing
irrelevant tools after providing the same hint.

\paragraph{Tools and inference.}
RAG, Table, and Graph expose search, read-only DuckDB, and graph-traversal
tools, respectively. We use a portable text-based ReAct loop capped at eight
tool turns (median four), with temperature 0 and greedy decoding. S4 and S6
expose all three surfaces, creating distractor pressure; the other settings
restrict tool access as defined above. Each task's common Efficiency budget is fixed at
$2\times$ the token count of the canonical GPT-4o-mini S5 trace. Exact setting and tool schemas appear in
Appendix~\ref{sec:agent-details}; implementation and cost details appear in
Appendix~\ref{sec:reproducibility}.

\subsection{Main Results}
\label{sec:mainresults}

% ---- end table_persurface.tex ----
Table~\ref{tab:main} reports component and aggregate scores for every
model--setting pair. Tool access generally improves Answer over No-tool, but
the gains are non-monotonic. Gemini attains the highest aggregate (78.9, S5);
among unconstrained S4 agents it also leads at 76.5, followed by DeepSeek
(69.4), GPT-5.5 (69.3), and GPT-4o-mini (57.9). Across all backbones, S4
improves aggregate over S3 by 14.4--29.5 points, yet remains 2.4--13.9 points
below each model's best gold-informed condition. Component scores reveal
distinct optima: Gemini's highest Answer occurs in S6 (78.8), although S5
gives a marginally higher aggregate; GPT-4o-mini and GPT-5.5 show the same
answer--aggregate trade-off. \textbf{No single setting dominates every
component.}

\subsection{Separating Surface Hints from Tool Restriction}
\label{sec:routing-answer}
The matched S4$\to$S6 contrast isolates the effect of naming the required
surfaces while keeping all tools available. Route~F1 changes by $+13.5$,
$+1.8$, $+4.5$, and $+3.6$ points for GPT-4o-mini, DeepSeek, Gemini, and
GPT-5.5; Answer changes by $+9.9$, $-2.2$, $+2.8$, and $+4.4$. Source-task
clustered 95\% intervals exclude zero for the three positive Answer changes,
but include zero for DeepSeek. \textbf{Surface hints help several models, but
their value is model-dependent rather than automatic.}

The S6$\to$S5 contrast then isolates removal of irrelevant tools under the
same hint. Route~F1 increases by 1.8--18.7 points and Efficiency by 8.8--22.3,
yet Answer changes by $-2.5$, $+5.0$, $-3.5$, and $-6.3$ points. Only the
DeepSeek gain and GPT-5.5 decline have clustered intervals excluding zero.
Thus, restricting tools makes traces cleaner and cheaper, but does not
uniformly improve the final answer.

\subsection{Performance by Surface Composition}
\label{sec:persurface}
Table~\ref{tab:persurface} and Figure~\ref{fig:main}c decompose Answer by task
type. RAG, Table, Graph, and Cross peak at 84.0\%, 91.0\%, 73.9\%, and 75.8\%,
respectively. Cross is therefore not uniformly the lowest category: Gemini
reaches 75.8 under S6. Nevertheless, each model's best cross-surface score is
13.0--45.7 points below its own best single-surface score. \textbf{Evidence
composition remains a substantial, but model-dependent, challenge.}
The profiles differ sharply. GPT-4o-mini is strongest on Table, DeepSeek pairs
strong Table performance with moderate Cross performance, and Gemini is much
more robust on Cross. GPT-5.5 retrieves competitive single-surface answers but
drops to 48.2 on Cross. These differences argue against describing the dataset
as uniformly difficult or balanced by outcome. Per-surface reporting is essential.

\subsection{Post-Routing Bottlenecks}
\label{sec:failures}
The gap between routing and execution is clearest under S5: all four models
obtain 98.7--99.8 Route~F1, but Evidence remains 66.7--78.5 and Answer
56.1--75.3. Low S3 Route and Evidence are consistent with selection failures;
high Route with incomplete Evidence under S5 indicates acquisition failures;
and residual Answer errors after strong Evidence indicate computation or
synthesis failures. Because Evidence records artifact access rather than
semantic use, these stages are diagnostic categories, not causal proofs.
\textbf{Near-perfect routing still leaves 24.7--43.9\% Answer error.}

Trace behavior supports this separation. Exact-set routing under S5 is
95.9--99.4\%, with no unavailable tool family exposed. Under S6, however,
7.8--63.5\% of tasks contain at least one unnecessary tool call, depending on
the model. The same hint can therefore yield very different execution policies.

\subsection{Closed-Book Answerability and Efficiency}
\label{sec:closedbook}
No-tool Answer peaks at 2.4\%, so S1 is a closed-book diagnostic rather than a
contamination test. Its 95.8\% mean Efficiency reflects short, evidence-free
traces rather than task success and must be interpreted jointly with Answer.
Under the common budget defined in \S\ref{sec:setup}, mean Efficiency is 37.1\%
for ReAct-all, 35.7\% for Gold-hint/all, and 50.2\% for Gold-constrained.
Surface hints alone do not reduce exploration: S4$\to$S6 changes Efficiency by
$-0.8$ to $-3.5$ points. Tool restriction does: S6$\to$S5 gains 8.8--22.3
points. \textbf{Efficiency gains come from reducing the action space, not
merely naming the destination.} Efficiency remains a normalized,
tokenizer-dependent diagnostic weighted at 10\%; provider-dependent costs are
discussed in Appendix~\ref{sec:reproducibility}.

\section{Conclusion}
\label{sec:conclusion}

WorkSurface-Bench separates routing from answering on 1{,}151 auditable tasks
over documents, tables, and dependency graphs. Across four backbones and six
settings, routing and Answer are only moderately associated ($\rho=0.62$).
Gold-constrained agents reach 98.7--99.8 Route F1, yet Answer remains
56.1--75.3. The matched S6 condition shows why the distinction matters:
surface hints can improve answers, while removing irrelevant tools more
reliably improves routing and efficiency. These interventions are controls,
not evidence of autonomous routing ability. The decomposed metrics localize
selection, acquisition, and downstream computation failures without treating
artifact access as proof of correct reasoning. Agent evaluations should
therefore report surface selection separately and test whether a system can
convert retrieved evidence into a correct answer. Future work should broaden
workspaces, domains, and repeated runs, and study more adaptive tool policies.
WorkSurface-Bench provides a reproducible basis for tracking these trade-offs
as workspace agents and interfaces evolve.

\bibliographystyle{plainnat}
\bibliography{custom}

\clearpage
\beginappendix
\section{Quality Audits and Construction Details}
\label{sec:appendix}

\subsection{Rubric-to-Task Conversion}
The 100 source tasks carry 1{,}850 rubrics across four Workspace-Bench rubric
types (Basic Evaluation: 450, Outcome Evaluation: 1{,}030, Process Evaluation:
339, Result Evaluation: 31). Deterministic and proof-carrying generators first
produce a 2{,}000-item candidate pool. GPT-5.5 retains 1{,}465 items passing
answerability, gold correctness, naturalness, atomicity, and leakage checks;
independent DeepSeek and Gemini audits then yield 1{,}151 items passing the
strict criteria under at least two of three models.

\subsection{Surface Coverage Validation}
Each routable surface must support executable gold construction and a
non-trivial task set. The final release contains 453 table-bearing items with
reproducible DuckDB results, 601 RAG-bearing items with verified source spans,
and 600 graph-bearing items whose dependency paths match source annotations.
These counts overlap because a cross-surface task contributes to every surface
it requires. All five personas and all three surfaces remain represented after
screening.

\subsection{Gold-Answer Audit}
Because we never take a gold answer from LLM free text, we can \emph{execute}
or \emph{re-verify} every gold value directly. We audited the whole set:
\textbf{(i)} all 453 table-bearing items have queries that execute against the
DuckDB registry and reproduce their gold results; \textbf{(ii)} all 601
RAG-bearing items have an exact evidence span; and \textbf{(iii)} all 600
graph-bearing items have valid nodes, paths, or complete neighbor sets.
This audit is fully automated and re-runnable from the released code.

\subsection{Independent Human Audit}
Three annotators independently reviewed the same stratified 200-item sample:
30 Graph, 40 Table, 30 RAG, 25 Graph+Table, 35 RAG+Graph, 25 RAG+Table, and all
15 three-surface tasks. By majority vote, all 200 items are natural,
answerable, dependent on every annotated surface, gold-correct, atomic and
unambiguous, and free of a leakage cue. Required-surface necessity and
atomicity are unanimous on 200/200 items; 192/200 items are unanimous across
all six dimensions. Pairwise agreement ranges from 98.7\% (naturalness) to
100\% (surface necessity and atomicity).

\subsection{Distribution}
\label{sec:distribution}
Figure~\ref{fig:distribution} shows the benchmark's cross-cutting
distributions. Cross-surface is the largest class (42\%), followed by Table
(24\%), RAG (19\%), and Graph (15\%); personas span five roles with the operational two dominating,
reflecting the Workspace-Bench-Lite source; answer types cover number, list,
and string. Derivation paths are intentionally uneven: filtering retains every
strictly verified item rather than enforcing equal quotas, while no single
path accounts for a majority of the release.

\subsection{Closed-Book Answerability Diagnostic}
\label{sec:appendix-closedbook}
The S1 no-tool setting reaches at most 2.4\% Answer over all 1{,}151 tasks. This
result indicates limited closed-book answerability, but it is not a formal
contamination test and does not rule out item-level memorization.

\subsection{Task Examples}
Two illustrative items (abbreviated for space):
\smallskip
\noindent\textbf{cross\_surface (LLM-assisted, gold via query execution).}
\emph{Question}: ``For workspace task 107: what is the total sales amount
across the Asia Pacific region?''
\emph{Gold}: \texttt{800511.75} (number).
\emph{Gold evidence}: [surface=rag, file=\texttt{t107\_\_report.md},
span=``Asia Pacific''; surface=table, table=\texttt{t107\_\_asia\_pacific\_orders},
query=\texttt{SELECT ROUND(SUM(sales),2) FROM ...}].
\emph{Required surfaces}: \{rag, table\}.

\begin{figure}[t]
\centering
\includegraphics[width=0.98\textwidth]{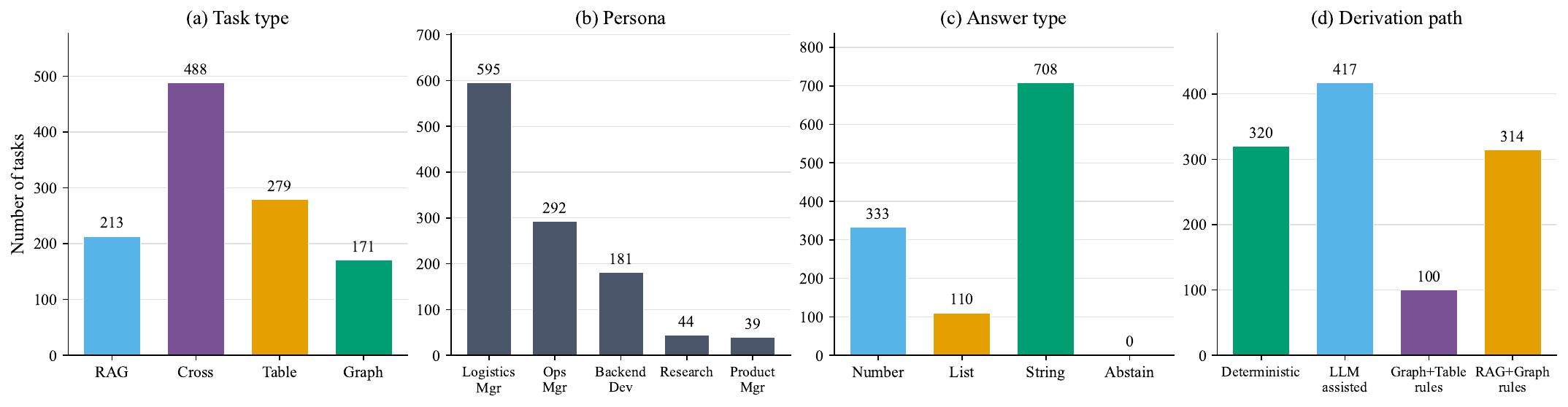}
\caption{Dataset distributions across (a) task type, (b) persona, (c) answer
type, and (d) derivation path. Frequencies are intentionally non-uniform:
filtering prioritizes verifiability and strict surface necessity over quotas.}
\label{fig:distribution}
\end{figure}

\section{Agent Configuration Details}
\label{sec:agent-details}

\begin{table}[t]
\centering\small
\begingroup
\renewcommand{\arraystretch}{1.10}
\setlength{\tabcolsep}{3.5pt}
\begin{tabular*}{\textwidth}{@{\extracolsep{\fill}}
>{\raggedright\arraybackslash}p{0.10\textwidth}
>{\raggedright\arraybackslash}p{0.15\textwidth}
>{\raggedright\arraybackslash}p{0.15\textwidth}
>{\raggedright\arraybackslash}p{0.21\textwidth}
>{\raggedright\arraybackslash}p{0.27\textwidth}}
\toprule
\textbf{Setting} & \textbf{Exposed surfaces} & \textbf{Initial behavior} & \textbf{Subsequent interaction} & \textbf{Purpose} \\
\midrule
\textbf{S1} No-tool & None & Answer directly & No tools & Closed-book answerability baseline \\
\cmidrule(lr){1-5}
\textbf{S2} Always-RAG & RAG only & Forced \texttt{kb\_search} & One-shot answer from returned snippets & Fixed single-surface baseline \\
\cmidrule(lr){1-5}
\textbf{S3} Naive-router & One selected surface & Choose exactly one of RAG/\allowbreak Table/\allowbreak Graph & ReAct restricted to that surface & One-shot surface routing \\
\cmidrule(lr){1-5}
\textbf{S4} ReAct-all & RAG, Table, Graph & No forced call & Dynamically interleave exposed tools, up to 8 turns & Unconstrained agent under distractor pressure \\
\cmidrule(lr){1-5}
\textbf{S5} Gold-constrained & Required surfaces only & No forced call & Dynamically interleave tools within the required set, up to 8 turns & Hint plus irrelevant-tool removal \\
\cmidrule(lr){1-5}
\textbf{S6} Gold-hint/all & RAG, Table, Graph & Required surfaces named in prompt & Dynamically interleave any exposed tools, up to 8 turns & Surface-information intervention \\
\bottomrule
\end{tabular*}
\endgroup
\caption{Operational definitions of the six agent settings. All settings use
the same question and answer format; only tool exposure and the forced initial
behavior differ.}
\label{tab:agent-settings}
\end{table}

Table~\ref{tab:agent-settings} gives the executable distinction among the six
settings. S5 and S6 receive only the names of the required surfaces: neither is
given the gold query, evidence, tool sequence, or answer. S5 restricts tool
exposure to those surfaces, whereas S6 retains all tools. Route scores are
computed from surfaces actually used in the trace, so neither condition is
assigned a perfect Route score by construction.

For S3--S6, each ReAct turn emits exactly one JSON object containing either a
tool call or \texttt{final\_answer}. RAG exposes \texttt{kb\_search}; Table
exposes \texttt{table\_list}, \texttt{table\_describe}, and read-only
\texttt{table\_query}; Graph exposes entity search, neighbor lookup, and
traversal. Invalid JSON receives a format-repair prompt but still consumes a
turn. Runs use greedy decoding at temperature 0 and terminate on a final
answer or after eight tool turns.

\subsection{Statistical and Scoring Robustness}
We report paired differences with a 2{,}000-replicate bootstrap clustered by
the originating Workspace-Bench task, so derived items from the same source
are resampled together. The intervals cited in \S\ref{sec:routing-answer} use
this procedure. We also recompute the aggregate under equal, Answer-heavy,
and Route-heavy weights. Component metrics and the S4--S6 intervention
conclusions are unchanged; some close aggregate rankings do change under equal
weights, which is why we treat Route, Evidence, Answer, and Efficiency as the
primary outcomes rather than the aggregate alone.

\section{Reproducibility, Cost, and Access}
\label{sec:reproducibility}

\paragraph{Software.}
We release, under an open-source license, (i) the source-download and
freeze scripts (Workspace-Bench-Lite commit hash + per-file SHA-256), (ii) the
surface-projection and SOP-metadata converters, (iii) the deterministic and LLM-assisted task
derivers with a shared verifiable-gold contract, (iv) the scoring package with
ten unit tests, and (v) the agent harness (portable text-based ReAct loop,
per-task backbone, task-level checkpointing).

\paragraph{Runs.}
Main experimental numbers come from one canonical trajectory per
(task, model, setting) through an OpenAI-compatible proxy. The initial sweep
used moderate concurrency; failed trajectories were retried at lower
concurrency, with successful retries replacing failed rows. The final release
contains 27{,}624 retained traces with no protocol errors for \texttt{gpt-4o-mini},
\texttt{deepseek-v4-pro}, \texttt{gemini-3.1-pro-preview}, and
\texttt{gpt-5.5}.

\paragraph{Cost.}
Exact monetary cost depends on the proxy and model pricing in effect at run
time. We therefore release per-trajectory token counts, retry status, and the
aggregation scripts, allowing costs to be recomputed under any provider's
pricing schedule.

\paragraph{License and access.}
The derived benchmark inherits Workspace-Bench-Lite's license (see the
source repository). Our derived artifacts (task JSONL, canonical KB text,
DuckDB parquet registry, surface graphs, scoring code, and agent harness) are
released under CC~BY~4.0 for data and Apache~2.0 for code. The
\texttt{wsb\_lock.json} freezes the source commit and per-file SHA-256 so any
future re-derivation is bit-reproducible.

\section{Datasheet}
\label{sec:datasheet}

We include an abbreviated Datasheet for Datasets~\cite{gebru2018datasheets}.

\paragraph{Motivation.}
WorkSurface-Bench was created to evaluate whether enterprise agents can
\emph{route} across heterogeneous knowledge surfaces (documents, tables,
dependency graphs), separately from answer correctness. Among the surveyed
benchmarks, none reports routing over these three workspace representations as
a separate metric.

\paragraph{Composition.}
Each instance is an atomic question over one persona-scoped workspace, with a
gold answer, required surfaces, expected tool types, and evidence annotations.
1{,}151 instances total; see Table~\ref{tab:stats-composition} and
Figure~\ref{fig:distribution} for distributions.

\paragraph{Collection process.}
Instances are derived automatically from Workspace-Bench-Lite (English split,
commit \texttt{60b08b1c}) via the pipeline in \S\ref{sec:dataset}. Three
human annotators independently audited a stratified 200-item sample; the
underlying WSB source inherits its authors' expert validation.

\paragraph{Preprocessing.}
Documents are rendered to canonical UTF-8 markdown; tables are normalized and
loaded into a per-workbook DuckDB view registry with provenance columns; the
dependency graph is enriched with programmatic edges. The raw source is
retained (frozen by hash), so preprocessing is fully reversible.

\paragraph{Uses.}
Intended: benchmarking enterprise routing agents, ablations over surface
selection, and reproducibility research.
Not intended: training data (the set is small and constructed from
mixed-provenance content), safety-critical decisions, or production
deployment gates.

\paragraph{Distribution and maintenance.}
The versioned dataset and code are hosted at the repository linked in the
abstract; issues accept fixes and error reports. Future releases will improve
persona coverage and held-out evaluation.

\paragraph{Ethics and privacy.}
The dataset contains no personally identifying or human-subject data. Files
are public resources or grounded synthetic artifacts inherited from
Workspace-Bench-Lite. Because some source content is LLM-generated,
memorization is possible; the closed-book diagnostic characterizes
answerability without claiming to rule out contamination.

\paragraph{Broader impact.}
The benchmark helps deployers test whether agents follow the right retrieval
path. Its main risk is over-generalization: scores on scenario-grounded hybrid
data should not serve as deployment gates without target-system evaluation.

\end{document}